\documentclass[times,twocolumn,final,authoryear]{elsarticle_modified}

\usepackage{ycviu_modified}

\usepackage{framed,multirow}

\usepackage{amssymb}
\usepackage{latexsym}
\usepackage{amsmath}
\usepackage{subcaption}
\usepackage{stackengine}
\usepackage{resizegather}
\usepackage{booktabs} 
\usepackage{lineno}
\usepackage{graphicx}
\usepackage{url}
\usepackage{xcolor}
\definecolor{newcolor}{rgb}{.8,.349,.1}

\journal{Computer Vision and Image Understanding}

\begin{document}

\begin{frontmatter}

\title{Cross-view image synthesis using geometry-guided conditional GANs}

% \author{Krishna Regmi\corref{cor1}} 

\author{Krishna \snm{Regmi}\corref{cor1}} 
\cortext[cor1]{Corresponding author: 
%   Tel.: +0-000-000-0000;  
%   fax: +0-000-000-0000;
}
\ead{krishna.regmi7@gmail.com}
\address{Center for Research in Computer Vision (CRCV), University of Central Florida, Orlando, FL, USA}
\author{Ali \snm{Borji}}
% \author[2]{Given-name \snm{Surname}}
\ead{aliborji@gmail.com}
\address{Markable.AI, Newyork, USA}

\begin{abstract}
We address the problem of generating images across two drastically different views, namely ground (street) and aerial (overhead) views. Image synthesis by itself is a very challenging computer vision task and is even more so when generation is conditioned on an image in another view. Due the difference in viewpoints, there is small overlapping field of view and little common content between these two views. Here, we try to preserve the pixel information between the views so that the generated image is a realistic representation of cross view input image. For this, we propose to use homography as a guide to map the images between the views based on the common field of view to preserve the details in the input image. We then use generative adversarial networks to inpaint the missing regions in the transformed image and add realism to it. Our exhaustive evaluation and model comparison demonstrate that utilizing geometry constraints adds fine details to the generated images and can be a better approach for cross view image synthesis than purely pixel based synthesis methods. 

\end{abstract}

\end{frontmatter}

{\let\thefootnote\relax\footnote{{This work extends our CVPR 2018 paper [\cite{Regmi_2018_CVPR}]}}}

\section{Introduction} 
Novel view synthesis is a long-standing problem in computer vision. Earlier works in this area synthesize views of single objects or natural scenes with small variation in viewing angle. 
%between the input and target images. 
For generating views of single objects in a uniform background or scenes, the task regards learning a mapping or transformation across views. 
%For natural scenes 
With a small camera movement, there is a high degree of overlap in field of views resulting in images with high content overlap. The generative models can learn to copy large parts of the image content from the input to the output and perform the synthesis task satisfactorily. Despite this, view synthesis task is very challenging due to presence of multiple objects in the scene. The network needs to learn the object relations and occlusions in the scene. 

Generating cross-view natural scenes conditioned on images from drastically different views (e.g., generating top-view from street view scene) is very painstaking. This is mainly because there is very little overlap between the corresponding field of views. Thus, simply copying and pasting pixels from one view to another would not be a solution. Rather, it is needed to learn the object classes present in the input view and to understand the correspondences in the target view with appropriate object relations and transformations (i.e., geometric reasoning). 

In this work, we address the problem of synthesizing ground-level images from overhead imagery and vice versa using conditional Generative Adversarial Networks [\cite{DBLP:journals/corr/MirzaO14}]. Also, when possible, we guide the generative networks by feeding homography transformed images as inputs to improve the synthesized results. Basically, conditional GANs try to generate new images from conditioning variables as input. The conditioning variables could be other images, text descriptions, class labels, etc. 
% Preliminary works in GANs utilize unsupervised learning to generate samples from latent representations or from a random noise vector [\cite{goodfellow2014generative}]. 

Our first approach exploits the success of the first GAN-based image-to-image translation network put forward by \cite{pix2pix2017} as a general purpose architecture on multiple image translation tasks. 
This work translates images of objects or scenes which are represented by RGB images, gradient fields, edge maps, aerial images, sketches, etc across these representations. Thus, it essentially operates on different representations of images in a single view. 
We use this architecture as a starting point (base model) in our task and obtain encouraging results. 
The limitation of this approach to our problem, however, is that the images to be transformed to are very different as they come from two drastically different views, have small fields of view overlap, and objects in the images might be occluded. As a result, learning to map the pixels between the views is difficult as the corresponding pixels in two views may represent different object classes. 
To address this challenge, here we propose to use the semantic segmentation maps of target view images to regularize the training process. This helps the network to learn the semantic class of the pixels in the target view and to guide the network to generate the target pixels. By encouraging the networks to generate the segmentation maps in target view, the network learns the semantic classes of each pixel which are important cues in assisting to generate cross-view images that preserve semantic information from the source view to the target view.     

The next approach we take to solve the cross-view image synthesis task is to exploit the geometric relation between the views to guide the synthesis. For this, we first compute the homography transformation matrix between the views and then project the aerial images to street-view perspective. By doing so, we obtain an intermediate image that looks very close to the target view image but not as realistic and with some missing regions. Now, our problem reduces to preserving the scene layout and details while filling in the missing regions and adding realism to the transformed image. For this, we use cGAN architectures described in previous approach. We also use different cGANs that work specifically for inpainting and realism tasks to preserve the pixel information from the homography transformed image in a controlled manner.

% % visible in the input view and simplify the learning process to synthesize the target view. The transformed image is not as realistic as the street-view image and also has missing regions for parts not visible in the input view (e.g., sky regions are missing in aerial-to-ground transformed image). can be inpainted and may not be as realistic as the streetview images. Next, we use GANs to inpaint missing regions and add realism to the image (i.e., fine tuning).

% , once the scene is transformed to the target perspective, enforcing the network to preserve the scene layout and details while synthesizing the missing regions and adding realism (i.e., finetuning the image) 

% from the input view in the target view as much as possible. The transformed images obtained in this way have missing regions and may not be as realistic as the streetview images. Next, we use GANs to inpaint missing regions and add realism to the image (i.e., fine tuning). 

% \begin{figure}
% \centering
% \includegraphics[width=0.45\textwidth]{Objective_v2.pdf}
% \vspace{-8pt}
% \caption{\small \label{fig:obj} Example images in overhead/aerial view (left) and ground-level/street-view (right). The images reflect the great diversity and richness of features in two views implying that the network needs to learn a lot for meaningful cross-view generation. We propose to use cGANs to solve this problem.}
% \vspace{-15pt}
% \end{figure} 

To summarize, we propose the following methods. We start with the simple image-to-image translation network of \cite{pix2pix2017} as a baseline (here called {\bf Pix2pix}). We then propose two new cGAN architectures that generate images as well as segmentation maps in the target view. Augmenting semantic segmentation generation to the architectures helps improve the quality of generated images. The first architecture, called {\bf X-Fork}, is a slight modification of the baseline, forking at the penultimate block to generate two outputs, target view image and segmentation map. The second architecture, called {\bf X-Seq}, has a sequence of two baseline networks connected. The target view image generated by the first network is fed to the second to generate its corresponding segmentation map. Once trained, both architectures are able to generate better images than the baseline that generates only the target view images. This implies that learning to generate segmentation map along with the image indeed improves the quality of generated images. We also use homography to transform the aerial image to ground view and feed the transformed image to these networks to further improve the results. Finally, we propose a method to preserve details from the homography transformed image in a controlled setting to generate the street view images. It constitutes two subtasks: a) generating missing regions by inpainting, and b) adding realism by using GAN to preserves details visible in aerial view into the street view image. We call this approach {\bf H-Regions}. Throughout the paper, {\bf H} in a method's name indicates that the input is the homography transformed image. 

% The rest of this paper is organized as follows. In Section \ref{sec:related_work}, we discuss the prior works related to ours. Next, we provide an overview of GAN in Section \ref{sec:background} and describe the proposed methods in Section \ref{sec:framework}. In Section \ref{sec:exp_setting}, we talk about the datasets and the experimental setups. We present the qualitative and quantitative results in Section \ref{sec:results} followed by discussions and conclusions in Section \ref{sec:conclusion}. 

\vspace{-5pt}
\section{Related Work}
\label{sec:related_work}

\subsection{Relating Aerial and Ground-level Images}
\cite{zhai2017crossview} explored the relationship between the cross-view images by learning to predict the semantic layout of the ground image from its corresponding aerial image. They used the predicted layout to synthesize ground-level panorama. Prior works relating the aerial and ground imageries have addressed problems such as cross-view co-localization [\cite{Lin_2013_CVPR, Vo2016}], ground-to-aerial geo-localization [\cite{DBLP:conf/cvpr/LinCBH15, mh2018cvm}] and geo-tagging the cross-view images [\cite{workman2015wide}]. Recently, the images generated by our cross-view image synthesis approach have been successfully used to bridge the domain gap between aerial and street-view images in geo-localization tasks [\cite{regmi2019bridging}].  

Cross-view relations have also been studied between egocentric (first person) and exocentric (surveillance or third-person) domains for different purposes. Human re-identification by matching viewers in top-view and egocentric cameras have been tackled by establishing the correspondences between the views in \cite{DBLP:conf/eccv/ArdeshirB16}. \cite{soran2014action} utilize the information from one egocentric camera and multiple exocentric cameras to solve the action recognition task. 
% \cite{DBLP:journals/corr/ArdeshirRB16} learn motion features of actions performed in ego- and exocentric domains to transfer motion information across the two domains. 

\vspace{-5pt}

\subsection{Learning View Transformations}
% \vspace{-5pt}
Existing works on viewpoint transformation have been conducted to synthesize novel views of the same objects [\cite{DTB17,10.1007/978-3-319-46478-7_20,10.1007/978-3-319-46493-0_18}]. \cite{10.1007/978-3-319-46493-0_18} proposed models that learn to copy the pixel information from the input view and utilize them to preserve the identity and structure of the objects to generate new views. \cite{10.1007/978-3-319-46478-7_20} trained an encode-decoder network to obtain 3D models of cars and chairs which they later used to generate different views of an unseen car or chair. \cite{DTB17} learned generative models by training on 3D renderings of cars, chairs and tables and synthesized intermediate views and objects by interpolating between views and models.

\vspace{-5pt}

\subsection{GAN and cGAN}
% \vspace{-5pt}
\cite{goodfellow2014generative} are the pioneers of Generative Adversarial Networks that are very successful at generating sharp and unblurred images, much better compared to existing methods such as Restricted Boltzmann Machines [\cite{Hinton:2006:FLA:1161603.1161605, Smolensky:1986:IPD:104279.104290}] or deep Boltzmann Machines [\cite{salakhutdinov2009deep}]. 

Conditional GANs synthesize images conditioned on different parameters during both training and testing. Examples include conditioning on labels of MNIST to generate digits by \cite{DBLP:journals/corr/MirzaO14}, on image representations to translate an image between different representations by \cite{pix2pix2017}, and generating panoramic ground-level scenes from aerial images of the same location by \cite{zhai2017crossview}. \cite{pmlr-v48-reed16} synthesize images conditioned on detailed textual descriptions of the objects in the scene, and \cite{han2017stackgan} improved on that by using a two-stage Stacked GAN. 

% [[look for few recent works and add here!!!]]
\vspace{-5pt}

\subsection{Cross-Domain Transformations using GANs}
\cite{pmlr-v70-kim17a} utilized the GAN networks to learn the relation between images in two different domains such that these learned relations can be transferred between the domains. Similar work by \cite{CycleGAN2017} learned mappings between unpaired images using cycle-consistency loss. They assume that a mapping from one domain to the other and back to the first should generate the original image. Both works exploited large unpaired datasets to learn the relation between domains and formulated the mapping task between images in different domains as a generation problem. \cite{CycleGAN2017} compare their generation task with previous works on paired datasets by \cite{pix2pix2017}. They conclude that the result with paired images is the upper-bound for their unpaired examples. 

\vspace{-5pt}

\subsection{Geometry-guided Synthesis}
\cite{songgeometry} propose geometry-guided adversarial networks to synthesize identity-preserving facial expressions. The facial geometry is used as a controlled input to guide the network to synthesize facial images with desired expressions.
Similar work by \cite{kossaifi2017gagan} improves the visual quality of synthesized images by enforcing a mechanism to control the shapes of the objects. They map the generator's output to a mean shape and implicitly enforce the geometry of the objects and also add skip connections to transfer priors to the generated objects.

\vspace{-5pt}

\subsection{Image Inpainting}
\cite{pathak2016context} generated missing parts of images using networks trained jointly with adversarial and reconstruction losses and produced sharp and coherent images. \cite{yeh2017semantic} tackle the problem of image inpainting by searching for the encoding of the corrupted image that is closest to another image in the latent space and passing it through the generator to reconstruct the image. The closeness is defined based on the weighted context loss of the corrupted image, and a prior loss that penalizes unrealistic images. \cite{yang2017high} propose a multi-scale patch synthesis approach for high-resolution image inpainting by jointly optimizing on image content and texture constraint.

%  [[Add a paragraph or section on using geometry information in GANs]]
%  e.g., https://arxiv.org/pdf/1802.01822.pdf
%  DualGAN: Unsupervised Dual Learning for Image-to-Image Translation
%  Z Yi, H Zhang, P Tan, M Gong
%  arXiv preprint arXiv:1704.02510

%  also see works by this guy:

%  https://scholar.google.com/citations?hl=en&user=q6Caj6sAAAAJ&view_op=list_works&sortby=pubdate

% You can add more visualizations since we have space here. or any thing else in the supplement can go here

% also add some works on inpainting. I have some refs in that REU project on inpainting 
%  ]]

\vspace{-5pt}
\section{Background on GANs}
\label{sec:background}
Th Generative Adversarial Network (GAN) proposed by \cite{goodfellow2014generative} consists of two adversarial networks: a generator and a discriminator that are trained simultaneously based on the min-max game theory. The generator $G$ is optimized to map a $d$-dimensional noise vector (usually $d$=100) to an image (i.e., synthesizing) that is close to the true data distribution. The discriminator $D$, on the other hand, is optimized to accurately distinguish between the synthesized images coming from the generator and the real images coming from the true data distribution. The objective function of such a network is 
\vspace{-15pt}

\begin{equation}
\begin{split}
\resizebox{0.8\hsize}{!}{$%
\stackanchor{min }{G}  \stackanchor{max  }{D} L_{GAN} (G,D) $ = $ E_{x} {\raise.01ex\hbox{$\scriptstyle\sim$}}_{p_{data}(x)} [log D(x) ]+
$%
}%
\\ \hspace*{0.5in}
\resizebox{0.48\hsize}{!}{$%
E_{z}  {\raise.01ex\hbox{$\scriptstyle\sim$}}_{p_z(z)}[log(1 - D(G(z)))],
$%
}%
\end{split}
\end{equation}

\noindent where $x$ is real data sampled from data distribution ${p_{data}}$ and $z$ is a $d$-dimensional noise vector sampled from a Gaussian distribution ${p_{z}}$. 

Conditional GANs synthesize images looking into some auxiliary variable which may be labels [\cite{DBLP:journals/corr/MirzaO14}], text embeddings [\cite{han2017stackgan,pmlr-v48-reed16}] or images [\cite{pix2pix2017,CycleGAN2017, pmlr-v70-kim17a}]. In conditional GANs, both the discriminator and the generator networks receive the conditioning variable represented by $c$ in Eqn.~\eqref{eq_cond}. The generator uses this additional information during image synthesis while the discriminator makes its decision by looking at the pair of conditioning variable and the image it receives. Real pair input to the discriminator consists of true image from distribution and its corresponding label while the fake pair consists of the synthesized image and the label. For conditional GAN, the objective function is 

\vspace{-15pt}

\begin{equation}\label{eq_cond}
\begin{split}
\resizebox{0.8\hsize}{!}{$%
\stackanchor{min }{G} \stackanchor{max  }{D} L_{cGAN}(G,D) = E_{x,c} {\raise.01ex\hbox{$\scriptstyle\sim$}}_{p_{data}(x, c)} [log D(x,c)] 
$%
}%
\\ \hspace*{0.5in}
\resizebox{0.6\hsize}{!}{$%
+ E_{x', c} {\raise.01ex\hbox{$\scriptstyle\sim$}}_{p_{data}(x',c)}[ log(1 - D(x',c))],
$%
}%
\end{split}
\end{equation}
where $x'$ = $G(z,c)$ is the generated image.

In addition to the GAN loss, previous works [e.g.,~\cite{pix2pix2017,CycleGAN2017,pathak2016context}] have tried to minimize the $L1$ or $L2$ distances between real and generated image pairs. This step aids the generator to synthesize images very similar to the ground truth. Minimizing the $L1$ distance generates less blurred images than minimizing the $L2$ distance. That is, using the $L1$ distance increases image sharpness in generation tasks. Therefore, we use the $L1$ distance in our method. The expression to minimize the $L1$ distance is
\vspace{-15pt}

\begin{equation}\label{eq_cond_l1}
\begin{split}
\resizebox{0.7\hsize}{!}{$%
\stackanchor{min  }{G} L_{L1}(G)=E_{x,x'} {\raise.01ex\hbox{$\scriptstyle\sim$}}_{p_{data}(x,x')}[\mid \mid x - x' \mid \mid _1],
$%
}%
\end{split}
\end{equation}

The objective function for such conditional GAN network is the sum of Eqns. \eqref{eq_cond} and~\eqref{eq_cond_l1}.

Considering the synthesis of the ground level imagery $(I_{g})$ from aerial image $(I_{a})$, the conditional GAN loss and $L1$ loss are represented as in Eqns. \eqref{eq_gan} and \eqref{eq_l1}, respectively. For ground to aerial synthesis, the roles of $I_a$ and $I_g$ are reversed.
\vspace{-15pt}

\begin{equation}\label{eq_gan}
\begin{split}
\resizebox{0.8\hsize}{!}{$%
\stackanchor{min }{G} \stackanchor{max  }{D} L_{cGAN}(G,D) = E_{I_{g},I_{a}} {\raise.01ex\hbox{$\scriptstyle\sim$}}_{p_{data}(I_{g},I_{a})} [log D(I_{g},I_{a})]
$%
}%
\\ \hspace*{0.5in}
\resizebox{0.6\hsize}{!}{$%
 + E_{I_a, I_g'} {\raise.01ex\hbox{$\scriptstyle\sim$}}_{p_{data}(I_{a}, I_g')}[ log(1 - D(I_g',I_{a}))],
$%
}%
\end{split}
\end{equation} 

\vspace{-10pt}

\begin{equation}\label{eq_l1}
\begin{split}
\resizebox{0.75\hsize}{!}{$%
 \stackanchor{min  }{G} L_{L1}(G)=E_{I_g,I_g'} {\raise.01ex\hbox{$\scriptstyle\sim$}}_{p_{data}(I_g,I_g')}[\mid \mid I_g - I_g' \mid \mid _1],
$%
}%
\end{split}
\end{equation}
where $I_g' = G(I_{a})$.
The objective function for an image-to-image translation network is the sum of conditional GAN loss in Eqn. \eqref{eq_gan} and $L1$ loss in Eqn. \eqref{eq_l1}, as represented in Eqn. \eqref{eq_comb}: 

\vspace{-15pt}

\begin{equation}\label{eq_comb}
\begin{split}
\resizebox{0.65\hsize}{!}{$%
 L_{network} = \lambda_1 L_{cGAN} (G,D) + \lambda_2 L_{L1}(G),
$%
}%
\end{split}
\end{equation}
where, $\lambda_1$ and $\lambda_2$ are the balancing factors between the losses.

\vspace{-5pt}
\section{Framework}
\label{sec:framework}
In this section, we discuss the baseline methods and the proposed architectures for the task of cross-view image synthesis. 
% In this section, we discuss about the proposed architectures along with the baselines for the task of cross-view image synthesis. 

\subsection{Baselines}

The naive way to approach this task is to consider it as an image to image translation problem. We run the experiments in the following settings as our baselines. 
% In order to fully analyze the possible input scenarios to the network, we explore the following as baselines in image-to-image formulation:

\vspace{-5pt}

\subsubsection{Cross-view Image-to-Image Translation (X-Pix2pix)}
For this, the generator is an encoder-decoder network that takes in an image in first view as input and learns to generate the image in the other view as output. 
% Such an idea has be explored in the conference paper this work is based on \cite{regmi2018cross} and seems to work quite well. We also use similar approach as one of the baselines in this work. 

\vspace{-5pt}

\subsubsection{Cross-view Image-to-Image Translation with Stacked Outputs (X-SO)} 
The network takes an image in first view as input and generates a single output of 6 channels, the first 3 channels correspond to the RGB image and the next three channels represent the segmentation map. The L1 loss and adversarial loss are computed over six channels of output and the corresponding ground truth images.
% .generates target view image and the segmentation map stacked together as a 6-channel image; first 3 channels for image and the next three channels for its segmentation map. The L1 loss and adversarial loss are computed over six channels of output and the corresponding ground truth images. 

\begin{figure}[t]
\begin{tabular}{c}
% \vspace{-10pt}
\subcaptionbox{X-Fork architecture. \label{fig:crossview-fork}}{\includegraphics[width=0.80\linewidth]{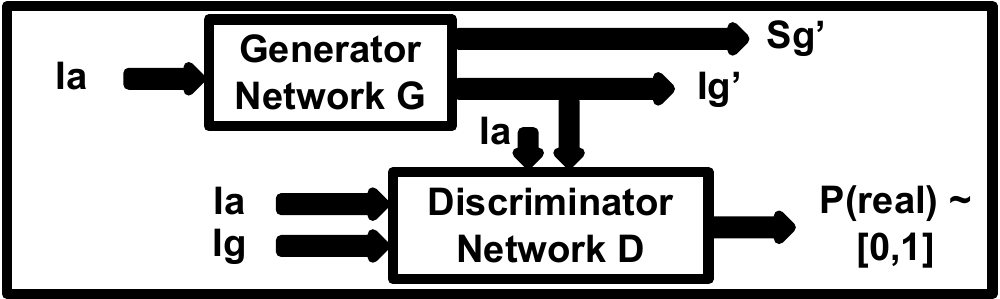}} \\
\subcaptionbox{X-Seq architecture. \label{fig:framework}}{\includegraphics[width=0.94\linewidth]{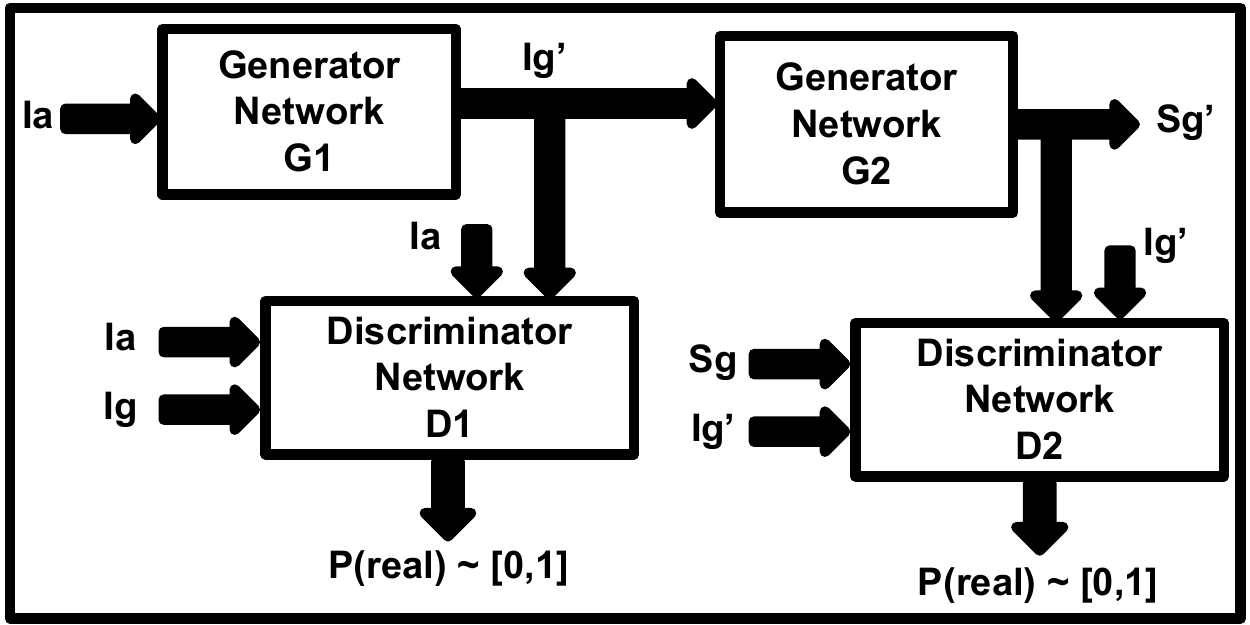}}
\end{tabular}
\vspace{-10pt}
\caption{\small Our proposed network architectures. a) X-Fork: Similar to baseline architecture except that G forks to synthesize image and segmentation map in target view, and b) X-Seq: a sequence of two cGANs, G1 synthesizes target view image that is used by G2 for segmentation map synthesis in corresponding view. In both architectures, $I_a$ and $I_g$ are real images in aerial and ground views, respectively. $S_g$ is the ground-truth segmentation map in street-view. $I_g'$ and $S_g'$ are synthesized image and segmentation map in ground view.}
 \vspace{-10pt}
\end{figure}

\subsection{Proposed Methods}
\subsubsection{\label{fork} Crossview Fork (X-Fork)}
\begin{figure}
\centering
\includegraphics[width=0.48\textwidth]{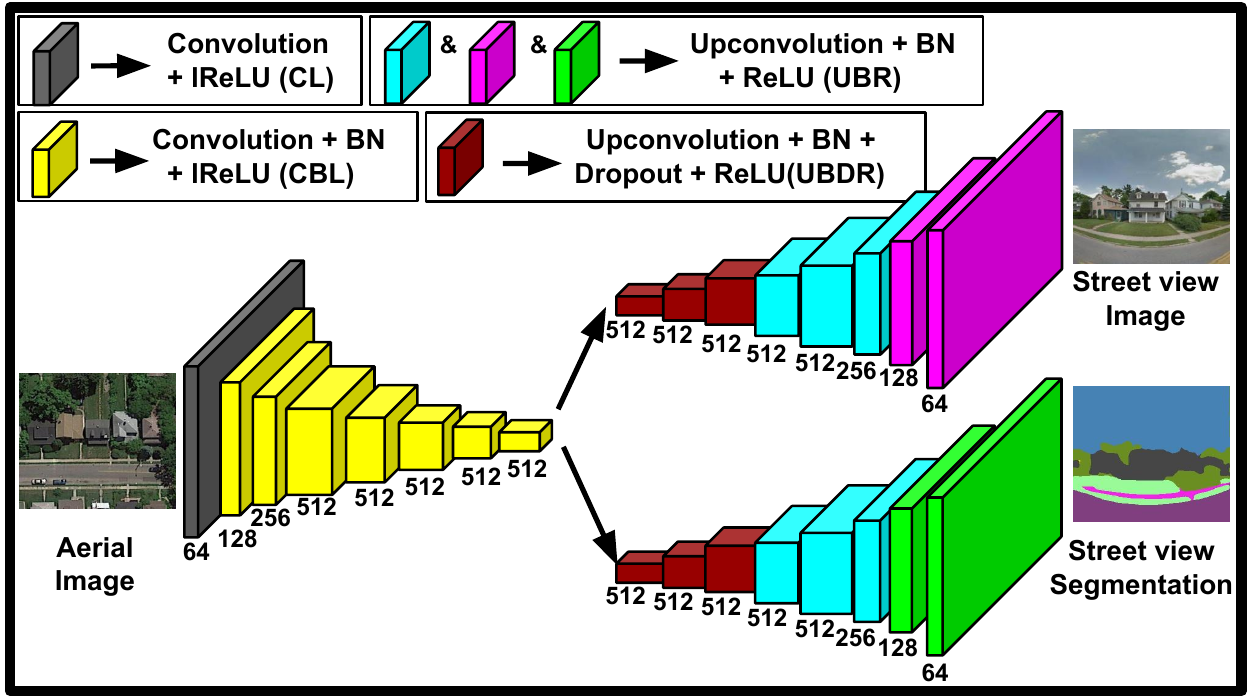}
% \vspace{-20pt}
\caption{\small \label{fig:crossview-fork-gen}Generator of X-Fork architecture in Figure \ref{fig:crossview-fork}. BN means batch-normalization layer.
The first six blocks of decoder share weights, forking at the penultimate block. 
The number of channels in each convolution layer is shown below each blocks. }
\vspace{-15pt}
\end{figure}

% \vspace{5pt}
Our first architecture, known as the Crossview Fork, is shown in Figure \ref{fig:crossview-fork}. The generator network is forked to synthesize two outputs of 3 channels each, the first output is the RGB image and the second output is the segmentation map both in target view. The fork-generator architecture is shown in Figure \ref{fig:crossview-fork-gen}. The first six blocks of decoder share the weights. This is because the image and segmentation map contain a lot of shared features. The number of kernels used in each layer (block) of the generator are shown below the blocks. 

The inherent idea behind this architecture is multi-task learning by the generator network. When the generator is enforced to learn the semantic class of the pixels together with the image synthesis, this helps to improve the image synthesis task. The generated segmentation map serves as an auxiliary output. The objective function for this network is shown in Eqn. \ref{eq_fork}. 

\vspace{-15pt}

\begin{equation}\label{eq_fork}
\begin{split}
\resizebox{0.8\hsize}{!}{$%
L_{X-Fork} = \lambda_1 L_{cGAN}(G, D) + \lambda_2 L_{L1}(G(I_g,I_{g'}))+ \lambda_2 L_{L1}(G(S_g,S_{g'}))
$%
}%
\end{split}
\end{equation}

Note that the L$_1$ loss is minimized for images as well as segmentation maps whereas the adversarial loss is optimized for images only. This is because we only care about pixel accuracy for segmentation maps.

\vspace{-5pt}

\subsubsection{\label{seq} Crossview Sequential (X-Seq)}
% \vspace{-5pt}
Our second architecture uses a sequence of two cGAN networks as shown in Figure \ref{fig:framework}. The first network performs cross-view image synthesis and the generated image is fed to the second network as a conditioning input to synthesize its corresponding segmentation map. This two-stage end-to-end learning of image and segmentation map synthesis produces improved image quality compared to the network without second stage. The joint objective function for X-Seq architecture is shown in Eqn. \ref{eq_seq} below.
\vspace{-15pt}

\begin{equation}\label{eq_seq}
\begin{split}
\resizebox{0.75\hsize}{!}{$%
L_{X-Seq} = L_{network}(G_1, D_1) + L_{network}(G_2, D_2)
$%
}%
\end{split}
\end{equation}

\vspace{-5pt}

\subsubsection{Cross-view Pix2pix with Homography (H-Pix2pix)}
Another approach that we take here is to feed the homography transformed image as an input to the translation network. 
Our hypothesis is that the majority of the scene from the first view is transformed into the second perspective using the homography and this should ease the synthesis task. The large missing regions in the transformed images are mostly related to sky and buildings.

\vspace{-5pt}

\subsubsection{Cross-view Stacked Outputs with Homography (H-SO)} 
The network takes the homography transformed image  as input and generates target view image and the segmentation map stacked together as a 6-channel output; first 3 channels for image and the next three channels for its segmentation map.

\vspace{-5pt}

\subsubsection{Cross-view Fork with Homography (H-Fork)}
Here, we use the homography transformed image I$_{ah}$ as input to the network architecture proposed in subsection \ref{fork}. The hypothesis behind this idea is that the use of transformed images as input should ease the cross-view synthesis task compared to synthesizing by feeding the aerial images I${_a}$ directly.

\vspace{-5pt}

\subsubsection{Crossview Sequence with Homography (H-Seq)}
In this setup, we feed the homography transformed image I$_{ah}$ rather than the original input image I$_{a}$ as input to the X-Seq network of subsection \ref{seq}.

\vspace{-5pt}

\subsubsection{Crossview Regions with Homography (H-Regions)}
% For the images in SVA dataset, the camera that captures the aerial view images are very close to the street and is successful at capturing fine details in the image. 
In this method, we attempt to preserve the structural details visible in the aerial view images and guide the network to transfer those details to the synthesized ground view images. For this, we use the homography transformed image as input to our method and solve the synthesis task in following subtasks:

% and tackle the problem by sim
% Once we obtain the homography transformed image from the input image, we simplify the synthesis problem into smaller tasks. 
% We adopt a three-stage training.

% In this approach, we attempt to learn the geometric transformation between the views and utilize this information to aid the cross-view image synthesis task. We attempt to transfer the structural details visible in aerial view by first applying Homography Transformation and then preserving those details in synthesized ground view images.

% For homography, we employ a naive way of manually picking four corresponding points in images from two views and then computing the homography matrix which we apply to all the images in the dataset. We were really surprised to find that this matrix was generalizable to all the images in the dataset.

\begin{figure}[!t]
\centering
\includegraphics[scale=.68]{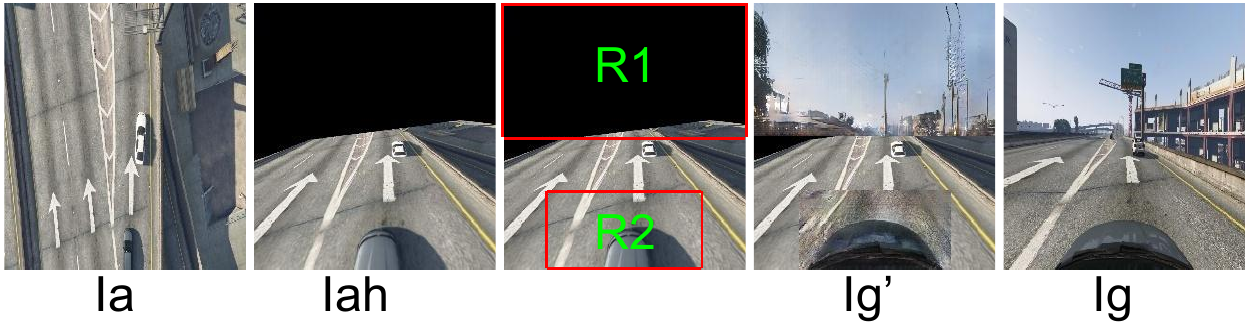}
% \vspace{-10pt}
\caption{\label{fig:regions} An aerial image I$_{a}$ shown in the left is first transformed to street-view perspective using homography (I$_{ah}$). The transformed image needs inpainting in upper box region (R${_1}$) and further processing in car region (R${_2}$). These two regions are generated and the region in between is copied from homography transformed image I$_{ah}$ to obtain I$_{g'}$. We further train I$_{g'}$ to smooth the region boundaries and add realism to it. I$_g$ is the corresponding ground truth image.
}
\vspace{-15pt}
\end{figure}

\textbf{Subtask I: }
The homography transformed image (I$_{ah}$ in Figure \ref{fig:regions}) has a large portion of missing region (R$_1$) in the image. Our first task is to fill in the missing region in the transformed images. We use an encoder-decoder network that takes I$_{ah}$ as input and generates only the upper half of the image (I$_{g'}$). 

\textbf{Subtask II: }
The street-view images are recorded by a camera mounted on a car's dashboard. Therefore, all the street-view images contain a part of the car's hood around the lower central region (R$_2$) in them which can be seen in image I$_{g}$ in Figure \ref{fig:regions}. Note that, this scenario is present in SVA dataset and can be avoided in real applications where the dash camera can be mounted such that the hood of the car is not captured in the frames. The homography transformed image I$_{ah}$ also has a car around that region which has been transformed from the aerial view of the car but it does not realistically represent a car in street-view. To address this, we mask a probable car-region and train a small network dedicated to learn mapping of the car region. This helps generate a realistic car region in ground view images (See region R$_2$ in image I$_{g'}$). 

Once we have images generated from the first two tasks, we copy them to their respective spatial locations in homography transformed image I$_{ah}$ creating a street-view image I$_{g'}$ and preserving the pixels of remaining regions.
% them together to get a complete image for which region 1 (upper half) comes from inpaint network, region 2 comes from car network and region 3 is copied from homography transformed image. 
Copying pixels in this manner helps us preserve the structural information that has been transformed using homography. The problem with this approach is that images do not look realistic at region boundaries. So, we further train another network to add realism to this image. 

\textbf{Subtask III: }
GANs are successful at generating images that look very realistic to human eyes. Here, we train a conditional GAN architecture on I$_{g'}$.
% We define our loss function such that we want to preserve the pixel information in the image while update around the region boundaries.
For this, we first define bands around the region boundaries as shown in Figure~\ref{fig:regions}. We formulate the loss function to preserve (by copying) the pixel information outside the bands to the output image while at the same time adding realism to the whole image. This step helps a lot to improve the visual quality of the synthesized image. 

We now define our loss functions for the subtasks.
For subtask one, we use conditional GAN network to inpaint missing regions in I$_{ah}$ by optimizing the network on adversarial and L${_1}$ losses for the missing regions only. For subtask two, we only consider region R$_2$ by masking out the remaining regions in input and output images and optimizing for adversarial and L${_1}$ losses for car region only. Once we have results from the above two subtasks, we compute I${_g'}$ as shown in Eqn. \ref{eq_regions}. 

\vspace{-15pt}

\begin{equation}\label{eq_regions}
\begin{split}
\resizebox{0.8\hsize}{!}{$%
I_{g'}$ = I$_{Inpaint}$ $\odot$ M$_1$ + I$_{car}$ $\odot$ M$_2$ + I$_{ah}$ $\odot$ (M - M$_1$ - M$_2$)
}%
\end{split}
\end{equation}

Here, I$_{Inpaint}$ is the image generated from the inpainting network of subtask one. I$_{car}$ is the car image generated in subtask two. M${_1}$ and M${_2}$ are 3-channel binary masks for regions R${_1}$ and R${_2}$, M being the 3-channel all ones image. The masks M${_1}$ and M${_2}$ are manually computed looking at the homography transformed image (I$_{ah}$) in Figure \ref{fig:regions}. This was done for a single frame only and worked well for all the images in the dataset. If the hood of the car was not visible in the street-view image, we wouldn't even need region R2 and correspondingly mask M${_2}$. $I_{g'}$ is fed to the realism network to generate the final image in the target view. $\odot$ is the element-wise product.

\vspace{-5pt}
\section{Experimental Setting}
\label{sec:exp_setting}
\vspace{-5pt}
\subsection{Datasets}
For the experiments in this work, we use three datasets described in the following.

\noindent \textbf{Dayton.} This cross-view image dataset is provided by \cite{Vo2016}. It consists of more than 1M pairs of street-view and overhead view images collected from 11 different US cities. We select 76,048 image pairs from Dayton images and create a train/test split of 55,000/21,048 pairs. We call it Dayton Dataset. The images in the original dataset have resolution of 354$\times$354. We resize them to 256$\times$256. We use this dataset for experiments in both aerial to ground \textbf{a2g} and ground to aerial \textbf{g2a} directions.

\noindent \textbf{CVUSA.}
We recruit this dataset [\cite{workman2015wide}] for direct comparison of our work with \cite{zhai2017crossview}. It consists of 35,532/8,884 train/test split of image pairs. Following \cite{zhai2017crossview}, the aerial images are center-cropped to 224 $\times$ 224 and then resized to 256 $\times$ 256. We only generate a single camera-angle image rather than the panorama. To do so, we take the first quarter of the ground level images as well as segmentations from the dataset and resize them to 256 $\times$ 256 in our experiments. 

\noindent \textbf{SVA.} The Surround Vehicle Awareness (SVA) dataset [\cite{10.1007/978-3-319-68560-1_21}] is a synthetic dataset collected from Grand Theft Auto V (GTAV) video game. The game camera is toggled between frontal and bird's eye view to simultaneously capture images in the two views at each game time step. We use the train/test split as provided in the dataset. The original dataset has 100 sets of training set images and 50 sets of test set images. The consecutive frames in each set are very similar to each other, so we use every tenth frame to remove redundancy in the dataset. Finally, we have a training set of 46,030 image pairs and a test set of 22,254 image pairs. The images are resized to 256 $\times$ 256 for experiments in this work. Sample images from SVA dataset are shown in the leftmost and rightmost columns of Figure \ref{fig:sva_qual}. We use this dataset for experiments in aerial to ground \textbf{a2g} direction only. Note that, homography related experiments are performed on this dataset only.

\begin{figure}
\centering
\includegraphics[width=0.45\textwidth]{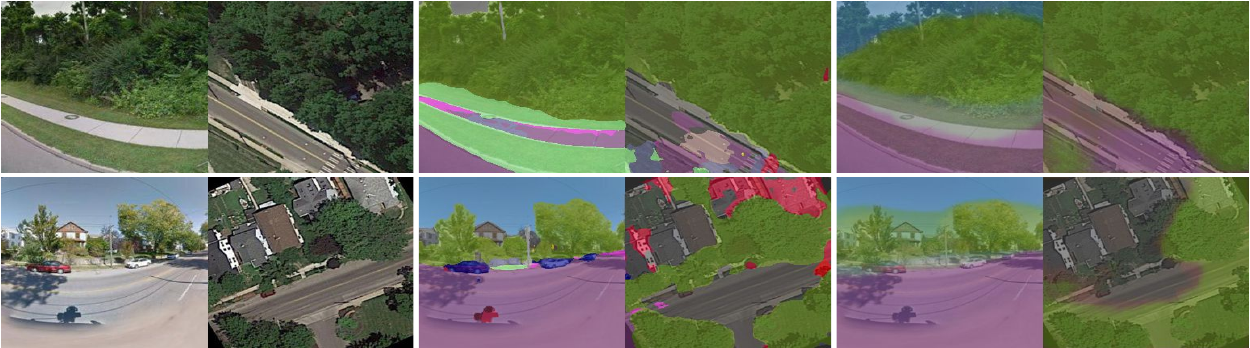}
\vspace{-5pt}
\caption{\small \label{fig:overlay}RGB image pairs from train set (left), segmentation masks from pre-trained RefineNet [\cite{Lin:2017:RefineNet}] overlaid on them (middle) and segmentation masks generated by X-Fork method overlaid on them (right).}
  \vspace{-15pt}
% \vspace{-17pt}
\end{figure}

The proposed Fork and Sequence networks learn to generate the target view images and segmentation maps conditioned on source view image or their homography transformed image. Training procedure requires the images as well as their semantic segmentation maps. The CVUSA dataset has annotated segmentation maps for ground view images, but for SVA and Dayton datasets such information is not available. To compensate, we use one of the leading semantic segmentation methods, known as the RefineNet [\cite{Lin:2017:RefineNet}]. This network is pre-trained on outdoor scenes of the Cityscapes dataset [\cite{Cordts2016Cityscapes}] and is used to generate the segmentation maps that are utilized as ground truth maps. These semantic maps have pixel labels from 20 classes (e.g., road, sidewalk, building, vegetation, sky, void, etc). Figure \ref{fig:overlay} shows image pairs from the Dayton dataset and their segmentation masks overlaid in both views. As can be seen, the segmentation mask (label) generation process is not perfect since it is unable to segment parts of buildings, roads, cars, etcetera in images.

\vspace{-5pt}
 
\subsection{Implementation Details} 

We use homography as a preprocessing step to transform the visual features from aerial images to ground perspective. Since the locations of aerial and ground view camera are fixed in the SVA dataset, first we randomly pick a pair of images from the dataset. We then manually select four points in the aerial image and find their corresponding locations in the ground-view image and use them to compute the homography matrix that transforms the aerial image to ground view and vice versa. Surprisingly, this method works well and avoids expensive computations for homography estimation usually done for each pair of images separately by computing SIFT features of the images and then finding the keypoints in the images. We also tried computing the SIFT features and then finding keypoints in two images and transforming images based on those points. This method could not find the corresponding points in two views, most likely because of very large perspective variation between the views.

We use the conditional GAN architecture very similar to \cite{pix2pix2017} as the base architecture. The generator is an encoder-decoder network with blocks of Convolution, Batch Normalization [\cite{Ioffe:2015:BNA:3045118.3045167}] and activation layers. Leaky ReLU with a slope of 0.2 is used as the activation function in the encoder, whereas the decoder has ReLU activation except for its final layer where Tanh is used. The first three blocks of the decoder have a Dropout layer in between Batch normalization and activation layer, with dropout rate of 50\%. The discriminator is taken as it is from the \cite{pix2pix2017}. For the experiments on the SVA dataset, we removed two blocks of CBL and UBDL from the generator architecture, primarily to save training time. We observed that removal of these blocks did not have much impact on the quality of synthesized images. Also, note that the synthesized semantic maps are 3-channel RGB images which effectively mitigated the class imbalance among the semantic classes. This was primarily done to consider all 20 semantic classes during the training and reduce bias towards dominant classes like houses, trees, road and sky. Had the semantic classes been limited to the dominant ones only, it would regularize the synthesized images to not learn the less prevalent objects in the target view images. Also, we brought in some confidence by the success of the pix2pix in synthesizing 3-channel segmentation maps from RGB images.

The convolutional kernels used in the networks are 4$\times$4 in size with a stride of 2. The upconvolution in the decoder is Torch [\cite{torch}] implementation of $SpatialFullConvolution$, and upsamples the input by 2. The convolutional operation downsamples the images by 2. No pooling operation is used in the networks. The $\lambda_1$ and $\lambda_2$ used in the objective function for different networks are the balancing factors between the GAN loss and the $L1$ loss. We fix $\lambda_1$ at 1 and $\lambda_2$ at 100 for Fork and Seq models. For realism task in the H-Regions method, $\lambda_1$=5 for adversarial loss and $\lambda_2$=2 for pixel-wise loss worked the best. Following the idea to smooth the labels by~\cite{DBLP:conf/cvpr/SzegedyVISW16} and demonstration of its effectiveness by \cite{DBLP:conf/nips/SalimansGZCRCC16}, we use one-sided label smoothing to stabilize the training process, replacing 1 by 0.9 for real labels. During the training, we utilized different data augmentation methods such as random jitter and horizontal flipping of images. The network is trained end-to-end with weights initialized using a random Gaussian distribution with zero mean and 0.02 standard deviation. Our methods are implemented in Torch [\cite{torch}]. Our code and data is available online at: 
https://github.com/kregmi/cross-view-image-synthesis.git.

We train the networks for 100 epochs on low resolution and 35 epochs on high resolution images of Dayton dataset and for 20 epochs on SVA dataset. Experiments are conducted on CVUSA dataset for comparison with the work of \cite{zhai2017crossview}. Following their setup, we train our architectures for 30 epochs using the Adam optimizer and moment parameters $\beta 1$ = 0.5 and $\beta 2$ = 0.999. We observed that the training for X-SO method on low resolution of Dayton dataset and CVUSA dataset was very unstable; so the qualitative and quantitative results are not very good for them. For H-Regions, we conduct experiments as follows. For subtask I, we train the network for 20 epochs. For subtask II, we train another network for one epoch only since the network needs to learn the mapping of the car and to preserve its color from source view to the target view. Eventually, we train a final realism network for 5 epochs (subtask III).

\vspace{-5pt}
\section{Evaluation and Model Comparison}
\label{sec:results}
We have conducted experiments in \textbf{a2g} (aerial-to-ground) and \textbf{g2a} (ground-to-aerial) directions on the Dayton dataset and \textbf{a2g} direction only on CVUSA and SVA datasets. 
We consider image resolutions of 64$\times$64 and 256$\times$256 on the Dayton dataset while for experiments on CVUSA and SVA datasets, 256$\times$256 resolution images are used. 

\vspace{-10pt}

\subsection{Evaluation measures} 
It is not straightforward to evaluate the quality of synthesized images [\cite{borji2018pros}]. In fact, evaluation of GAN methods continues to be an open problem [\cite{Theis2016a}]. Here we utilize four quantitative measures and one qualitative measure to evaluate our methods. 

\vspace{-5pt}

\subsubsection{Qualitative measure}
%\begin{itemize}
%User Study: 
For the subjective evaluation of different methods, we run a user study on the images synthesized using these methods. We show an aerial image along with the corresponding images in ground view synthesized using seven different methods to 10 users. We specifically ask each user to select the most realistic image that also contains the most visual details from the aerial view image. 
%\end{itemize}

\vspace{-5pt}

\subsubsection{Quantitative measures}
%\begin{itemize}
%\item 
\noindent \textbf{Inception Score:}
% \vspace{-5pt}
A common quantitative measure used in GAN evaluation is the \textit{Inception Score}
proposed by \cite{DBLP:conf/nips/SalimansGZCRCC16}. 
The core idea behind the inception score is to assess how diverse the generated samples are within a class while being meaningfully representative of the class at the same time. One major criticism regarding the Inception score is that the CNN trained on ImageNet objects may not be adequate for other scene datasets (our case). To address this, here we use the AlexNet model [\cite{DBLP:journals/cacm/KrizhevskySH17}] trained on Places dataset [\cite{zhou2017places}] with 365 categories to compute the inception score. The Places dataset has images similar to those in our datasets. 

We observe that the confidence scores predicted by the pre-trained model on our datasets are dispersed between classes for many samples and not all the categories are represented by the images. Therefore, we compute inception scores on Top-1 and Top-5 classes, where "Top-k" means that top k predictions for each image are unchanged while the remaining predictions are smoothed by an epsilon equal to (1 - $\sum$(top-k predictions))/(n-k classes). 

\noindent \textbf{Accuracy:} In addition to inception score, we compute the top-k prediction accuracy between real and generated images. We use the same pre-trained Alexnet model to obtain annotations for real images and class predictions for generated images. We compute top-1 and top-5 accuracies. For each setting, accuracies are computed in two ways: 1) considering all images, and 2) considering real images whose top-1 (highest) prediction is greater than 0.5. Below each accuracy heading, the first column considers all images whereas the second column computes accuracies the second way. 

\noindent \textbf{KL(model $\|$ data):} 
% \vspace{-5pt}
We compute the KL divergence between the model generated images and the real data distribution for quantitative analysis of our work, similar to some generative works [\cite{che-2016-reg-gan, nguyen2017dual}]. We again use the same pre-trained Alexnet described earlier. The lower KL score implies that the generated samples are closer to the real data distribution. 

\noindent \textbf{SSIM, PSNR and Sharpness Difference:}
We employ three measures from the image quality assessment literature to evaluate our models, similar to ~\cite{DBLP:journals/corr/MathieuCL15, DBLP:conf/cvpr/LedigTHCCAATTWS17,DBLP:conf/cvpr/ShiCHTABRW16,DBLP:journals/corr/ParkYYCB17}. Structural-Similarity (SSIM) measures the similarity between the images based on their luminance, contrast and structural aspects. SSIM value ranges between -1 and +1. Peak Signal-to-Noise Ratio (PSNR) measures the peak signal-to-noise ratio between two images to assess the quality of a transformed (generated) image compared to its original version. Sharpness difference (SD) measures the loss of sharpness during image generation. For each of these score, the higher the the better. Please refer to \cite{Regmi_2018_CVPR} for details on how to compute these scores.

\noindent \textbf{FID Score:} An alternative metric to evaluate the quality of generated images is by computing Frechet Inception Distance [\cite{NIPS2017_7240}] between the generated samples and the real images. We use the same AlexNet model (as above) pretrained on the Places Dataset to compute the FID score. The lower value of FID score for a method, the better. The FID scores that we obtain in this work are relatively larger than numbers reported in other works mainly because of the variations in the image statistics of the Places Dataset used during the training and our test images.
%\end{itemize}

% A common evaluation method is to show the generated images to human observers and ask their opinion about the images. Human judgment is based on the response to the question: Is this image (image-pair) real or fake? Alternatively, the images generated by different generative models can be pitted against each other and the observer is asked to select the image that looks more real.
% But in experiments involving natural scenes, such evaluation methods are more challenging as multiple factors often affect the quality of the generated images. For example, the observer may not be sure whether to base his judgment on better visual quality, higher sharpness at object boundaries, or more semantic information present in the image (e.g., multiple objects in the images, more details on objects, etc). 

\vspace{-10pt}

\subsection{Model Comparison} 
We conduct the qualitative and quantitative evaluation on 3 datasets. We report homography results on SVA dataset only.
% since the camera for aerial view images are closer to the ground and the images contain details which can be transformed to street view and are already very representative of ground view. Or basically, 
This is because the aerial view image for SVA dataset contains high overlap with the field of view of street-view image and thus application of homoraphy to preserve the details from aerial image seemed valid for this dataset compared to the other two. 

\begin{figure}
\centering
\includegraphics[width=0.48\textwidth]{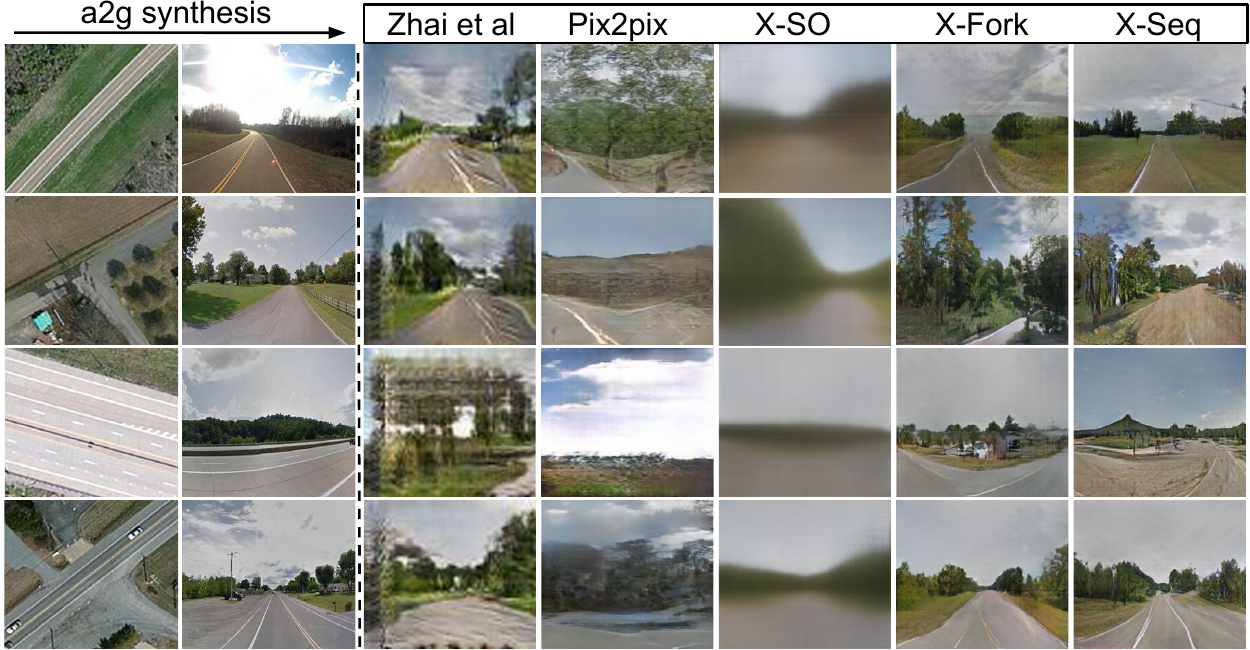}
\vspace{-18pt}
\caption{\small \label{fig:cvusa}Qualitative results of our methods and baselines on CVUSA dataset in \textbf{a2g} direction. First two columns show true image pairs, next four columns show images generated by \cite{zhai2017crossview}, X-Pix2pix [\cite{pix2pix2017}], X-SO, X-Fork and X-Seq methods respectively.}
% \vspace{-32pt}
%   \vspace{-15pt}
\end{figure}

We conduct the user studies to compare the images synthesized using different methods as well as illustrate qualitative results in Figures \ref{fig:test_64}, \ref{fig:cvusa} and \ref{fig:sva_qual} and conduct an in-depth quantitative evaluation on test images of the datasets.

\vspace{-5pt}

\subsubsection{Qualitative Comparison}

% \noindent \textbf{Results using SSIM, PSNR, and SD.}

\begin{figure*}
\centering
\includegraphics[width=0.49\textwidth]{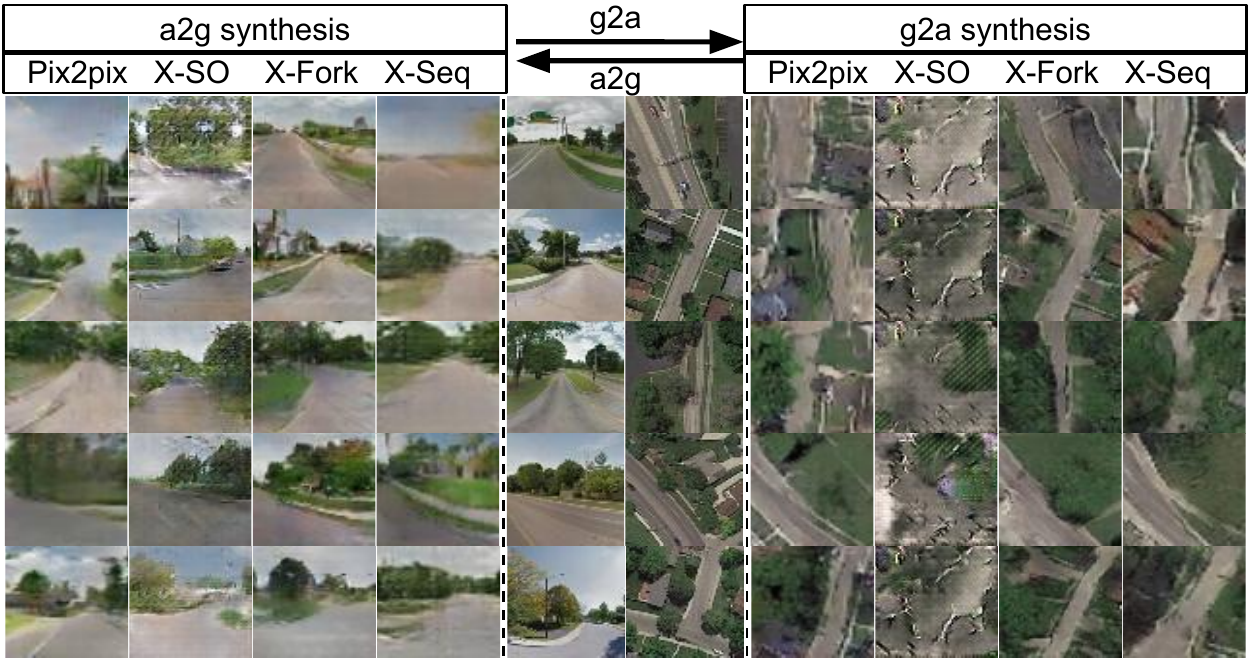} \hspace*{5pt}
\includegraphics[width=0.49\textwidth]{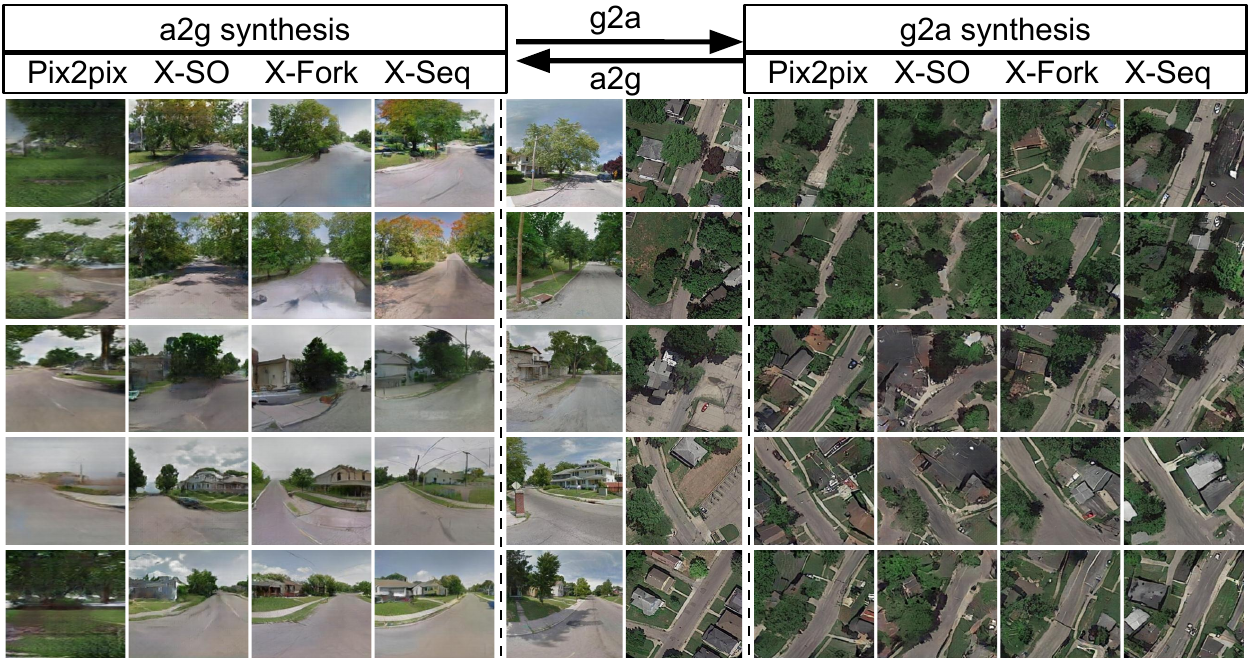}
\vspace{-18pt}
\caption{\small \label{fig:test_64}Example images generated by different methods in low (64 $ \times$ 64) resolution (left) 
and high (256 $ \times$ 256) resolution (right) in  \textbf{a2g} and \textbf{g2a} directions on the \textbf{Dayton} dataset.}
  \vspace{-10pt}
\end{figure*}

% We conduct experiments on all proposed methods as well as the baselines and illustrate the qualitative results on Dayton, CVUSA and SVA datasets.
% % in Figures \ref{fig:test_256}, \ref{fig:cvusa} and (reference to fig of sva dataset) respectively.
% For Dayton dataset, we observe that the generated images contain more details of objects in both views and the networks have utilized the knowledge of segmentation class to generate the objects at right pixels. Houses, trees, pedestrian lanes, and roads look very natural. Test results on CVUSA dataset show that images generated by proposed methods are visually better compared to Zhai \textit{et al.} \cite{zhai2017crossview} and the baselines. Results on SVA dataset show that our proposed homography transformation method, X-Hg is able to preserve the details in roads and road boundaries compared to the baselines. 
\noindent \textbf{Dayton dataset.} For 64$\times$64 resolution experiments, the networks are modified by removing the last two blocks of CBL from discriminator and encoder, and the first two blocks of UBDR from decoder of the generator. We run experiments on all three methods. Qualitative results are depicted in Figure \ref{fig:test_64} (left). The results affirm that the networks have learned to transfer the image representations across the views. Generated ground level images clearly show details about road, trees, sky, clouds, and pedestrian lanes. Trees, grass, road, house roofs are well rendered in the synthesized aerial images.

For 256$\times$256 resolution synthesis, we conduct experiments on all three architectures and illustrate the qualitative results in Figure \ref{fig:test_64} (right). We observe that the images generated in high resolution contain more details of objects in both views and are less granulated than those in low resolution. Houses, trees, pedestrian lanes, and roads look more natural.

\noindent \textbf{CVUSA dataset.} 
Test results on the CVUSA dataset show that images generated by the proposed methods are visually better compared to \cite{zhai2017crossview} and \cite{pix2pix2017}. Our proposed methods are more successful in transforming the pixel classes and thus generating correct class objects in target view compared to the baselines.

\begin{figure*}
\centering
\includegraphics[width=0.98\textwidth]{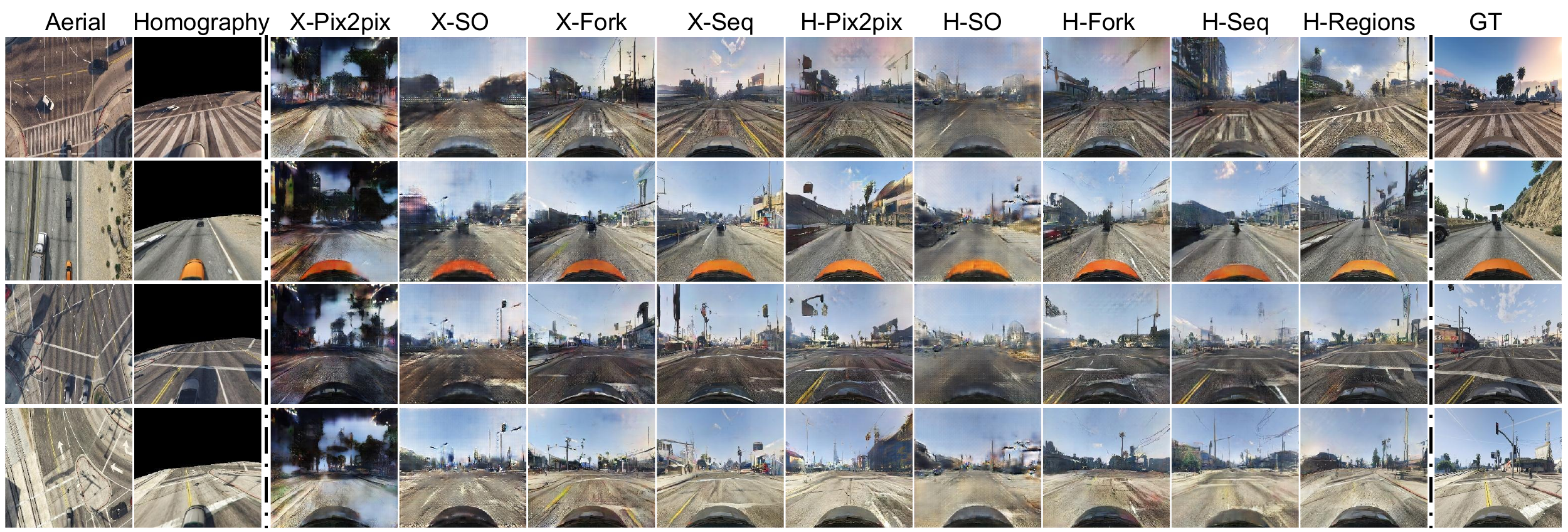}
\vspace{-5pt}
\caption{\small \label{fig:sva_qual} Example images generated by different methods in \textbf{a2g} direction for \textbf{SVA} dataset. }
% \vspace{-10pt}
\end{figure*}

\noindent \textbf{SVA dataset.}
The qualitative results on the SVA dataset for aerial to street-view synthesis is shown in Figure \ref{fig:sva_qual}. The proposed methods are capable at generating roads, car-hood, markers on road, sky and other details in the images. We observe that the images generated by the proposed H-Regions method contain more details around the central regions. This is due to enforcing the network to preserve those details from the aerial images.

\noindent \textbf{User Study.}
We conducted the perceptual test over 100 test images on 10 subjects to compare the images synthesized by different methods on the SVA dataset. The results are presented in Table \ref{tab:user_study}. The most preferred method is H-Regions, closely contested by H-Seq and X-Seq methods. The results illustrate the following: a) The use of homography transformed input drastically outperforms corresponding experiments with untransformed aerial image as input, and b) Users preferred the images synthesized using H-Regions because of the method's ability to preserve the pixel information onto the target view.

\begin{table}[t]
 \small
  \centering
  \renewcommand{\arraystretch}{.9}
  \renewcommand{\tabcolsep}{1mm}  
  
  \caption{\small \% of user preferences over images synthesized by different methods (over 100 images from the SVA dataset).}
  \vspace{-10pt}
  \label{tab:user_study}
    \begin{tabular}{ccc|ccc|c}
        \toprule
       \small X-Pix2pix & {X-Fork} & {X-Seq} & {H-Pix2pix} & {H-Fork}& {H-Seq}& {H-Reg.}\\
        \midrule
4 & 7.8 & 16.8 & 12.6 & 14.2 & 18.2 & 26.4 \\
        \bottomrule
    \end{tabular}
      \vspace{-15pt}
\end{table}

\vspace{-5pt}

\subsubsection{Quantitative Comparison}

\begin{table*}[t]
\small
  \renewcommand{\arraystretch}{.7}
  \centering
  \caption{\small Inception Scores of models over {\bf Dayton} and {\bf CVUSA} datasets.}
  \vspace{-10pt}
  \label{tab:inception_score}
    \begin{tabular*}{\textwidth}{l @{\extracolsep{\fill}} lccccccccc}
        \toprule
              \multicolumn{1}{l}{\textbf{Dir.}} & \multicolumn{1}{l}{\textbf{Methods}} & \multicolumn{3}{c}{\textbf{Dayton (64 $\times$ 64)}} & \multicolumn{3}{c}{\textbf{Dayton (256 $\times$ 256)}} & \multicolumn{3}{c}{\textbf{CVUSA}}     \\
                \cmidrule(lr){3-5}
\cmidrule(lr){6-8} \cmidrule(lr){9-11}
       $\rightleftarrows$  & & all  & Top-1  & Top-5  & all  & Top-1  & Top-5  & all  & Top-1  & Top-5  \\
         & &  classes &  class &  classes &  classes &  class &  classes &  classes &  class &  classes \\
        \midrule
      &  \cite{zhai2017crossview} & \textbf{--} & \textbf{--} & \textbf{--} & \textbf{--} & \textbf{--} & \textbf{--} & 1.8434 & 1.5171 & 1.8666\\
      &  X-Pix2pix  & 1.8029 & 1.5014 & 1.9300 & 2.8515 & 1.9342 & 2.9083 & 3.2771 & 2.2219 & 3.4312 \\
   a2g &  X-SO & {1.8577} & {1.4975} & {2.0301} & {2.9459} & {2.0963} & {2.9980} & 1.7575 & 1.4145 & 1.7791 \\
    &  X-Fork & \textbf{1.9600} & \textbf{1.5908} & \textbf{2.0348} & \textbf{3.0720} & \textbf{2.2402} & \textbf{3.0932} & 3.4432 & 2.5447 & 3.5567\\
   &  X-Seq & 1.8503 & 1.4850 & 1.9623 & 2.7384 & 2.1304 & 2.7674 & \textbf{3.8151} & \textbf{2.6738} & \textbf{4.0077}\\
    \cmidrule(lr){2-11}
   &  Real Data & 2.2096 & 1.6961 & 2.3008 & 3.7090 & 2.5590 & 3.7900 & 4.9971 & 3.4122 & 5.1150\\
     \midrule
  &  X-Pix2pix  & 1.7970 & 1.3029 & 1.6101 & 3.5676 & 2.0325 & 2.8141 & \textbf{--} & \textbf{--} & \textbf{--} \\
   &   X-SO & 1.4798 & 1.2163 & 1.5042 & 3.3397 & 2.0232 & \textbf{3.3485} & \textbf{--} & \textbf{--} & \textbf{--}\\
 g2a  &   X-Fork & \textbf{1.8557} & {1.3162} & \textbf{1.6521} & 3.1342 & 1.8656 & 2.5599 & \textbf{--} & \textbf{--} & \textbf{--}\\
  &   X-Seq & 1.7854 & \textbf{1.3189} & 1.6219 & \textbf{3.5849} & \textbf{2.0489} & {2.8414} & \textbf{--} & \textbf{--} & \textbf{--}\\
    \cmidrule(lr){2-11}
  &   Real Data & 2.1408 & 1.4302 & 1.8606 & 3.8979 & 2.3146 & 3.1682 & \textbf{--} & \textbf{--} & \textbf{--}\\
        \bottomrule
    \end{tabular*}
    %   \vspace{-10pt}
\end{table*}

\begin{table*}[t]
\small
  \renewcommand{\arraystretch}{.7}
  \centering
  \caption{\small Accuracies: Top-1 and Top-5 on {\bf Dayton} and {\bf CVUSA} datasets.}
  \vspace{-10pt}
  \label{tab:accuracies}
    \begin{tabular*}{\textwidth}{l @{\extracolsep{\fill}} llccccccccccc}
        \toprule
               \multicolumn{1}{l}{\textbf{Dir.}} & \multicolumn{1}{l}{\textbf{Methods}} & \multicolumn{4}{c}{\textbf{Dayton (64 $\times$ 64)}} & \multicolumn{4}{c}{\textbf{Dayton (256 $\times$ 256)}} & \multicolumn{4}{c}{\textbf{CVUSA}}     \\
                \cmidrule(lr){3-6}
\cmidrule(lr){7-10} \cmidrule(lr){11-14}
     $\rightleftarrows$   &  & \multicolumn{2}{c}{Top-1} & \multicolumn{2}{c}{Top-5} & \multicolumn{2}{c}{Top-1} & \multicolumn{2}{c}{Top-5} & \multicolumn{2}{c}{Top-1} & \multicolumn{2}{c}{Top-5} \\
        &  & \multicolumn{2}{c}{Accuracy (\%)} & \multicolumn{2}{c}{Accuracy (\%)} & \multicolumn{2}{c}{Accuracy (\%)} & \multicolumn{2}{c}{Accuracy (\%)} & \multicolumn{2}{c}{Accuracy (\%)} & \multicolumn{2}{c}{Accuracy (\%)} \\
          \midrule
      &   \cite{zhai2017crossview} & \textbf{--} & \textbf{--} & \textbf{--} & \textbf{--} & \textbf{--} & \textbf{--} & \textbf{--} & \textbf{--} & 13.97 & 14.03 & 42.09 & 52.29 \\
      &  X-Pix2pix  & 7.90 & 15.33 & 27.61 & 39.07 & 6.8 & 9.15 & 23.55 & 27.00 & 7.33 & 9.25 & 25.81 & 32.67 \\
a2g  &  X-SO & 4.68 & 7.49 & 16.72 & 24.96 & 27.56 & 41.15 & 57.96 & 73.20 & 0.29 & 0.21 & 6.14 & 9.08 \\
  &  X-Fork & \textbf{16.63} & \textbf{34.73} & \textbf{46.35} & \textbf{70.01} & 30.00 & 48.68 & 61.57 & 78.84 & \textbf{20.58} & \textbf{31.24} & \textbf{50.51} & \textbf{63.66}\\
  &  X-Seq & 4.83 & 5.56 & 19.55 & 24.96 & \textbf{30.16} & \textbf{49.85} & \textbf{62.59} & \textbf{80.70} & 15.98 & 24.14 & 42.91 & 54.41 \\
    \midrule
    &    X-Pix2pix  & 1.65 & 2.24 & 7.49 & 12.68 & 10.23 & 16.02 & 30.90 & 40.49 & \textbf{--} & \textbf{--} & \textbf{--} & \textbf{--} \\
    g2a  &  X-SO & 0.10 & 0.00 &  0.98 & 0.28 & 8.52 & 12.57 & 27.35 & 32.76 &\textbf{--} & \textbf{--} & \textbf{--} & \textbf{--}\\
 & X-Fork & \textbf{4.00} & \textbf{16.41} & \textbf{15.42} & \textbf{35.82} & 10.54 & 15.29 & 30.76 & 37.32 & \textbf{--} & \textbf{--} & \textbf{--} & \textbf{--}\\
  &  X-Seq & 1.55 & 2.99 & 6.27 & 8.96 & \textbf{12.30} & \textbf{19.62} & \textbf{35.95} & \textbf{45.94} & \textbf{--} & \textbf{--} & \textbf{--} & \textbf{--} \\
        \bottomrule
%         \vspace{-15pt}
    \end{tabular*}
      \vspace{-10pt}
\end{table*}

% \vspace{-10pt}

We report the quantitative results on three datasets next. 

\noindent \textbf{Dayton dataset.} The quantitative scores over different measures are provided in Tables \ref{tab:inception_score}, \ref{tab:accuracies}, \ref{tab:model_data_KL}, \ref{tab:fid} and \ref{tab:ssim_psnr_sd} under the columns Dayton (64 x 64) and Dayton (256 x 256) for low and high resolution synthesis.

Inception Score: The scores for X-Fork generated images are closest to that of real data distribution for Dayton dataset in low resolution in both directions and also in high resolution in \textbf{a2g} direction. The X-Seq method works best for \textbf{g2a} synthesis in high resolution for Dayton dataset. Inception scores on top-k classes follow a similar pattern as in all classes (except for Top-1 class on low resolution and Top-5 class on high resolution in \textbf{g2a} direction over Dayton dataset).

Accuracy: Results are shown in Table \ref{tab:accuracies}. We observe that the X-Fork method works better at low resolution whereas X-Seq is better at high resolution synthesis in both directions.

KL(model $\|$ data):
The scores are provided in Table \ref{tab:model_data_KL}. As it can be seen, our proposed methods generate much better results than the baselines. X-Fork generates images very similar to real distribution in all experiments except on the high resolution \textbf{a2g} experiment where X-Seq is slightly better than X-Fork. 

{FID Score:} The FID scores are presented in Table \ref{tab:fid}. X-Fork performs the best on lower resolution images while X-Seq works the best on higher resolution images in both directions \textbf{a2g} and \textbf{g2a}.  
% best for X-Fork method on 

{SSIM, PSNR, and SD:}
The scores are reported in Table \ref{tab:ssim_psnr_sd}. The X-Seq model works the best in \textbf{a2g} direction while X-Fork outperforms the rest in the \textbf{g2a} direction.

\noindent \textbf{CVUSA dataset.} The quantitative evaluation is shown in Tables \ref{tab:inception_score}, \ref{tab:accuracies}, \ref{tab:model_data_KL} and \ref{tab:ssim_psnr_sd} under the column CVUSA.

Inception Score: The X-Seq method works best for CVUSA dataset in terms of inception score.

Accuracy: Results are shown in Table \ref{tab:accuracies}. Images generated with X-Fork method obtain the best accuracy closely followed by X-Seq method.

KL(model $\|$ data):
The scores are provided in Table \ref{tab:model_data_KL}. As it can be seen, our proposed methods generate much better results than the baseline. X-Fork generates images very similar to real distribution and X-Seq very close to it. 

{FID Score:} X-Seq works the best on the CVUSA dataset closely followed by X-Fork network. See Table \ref{tab:fid}.

{SSIM, PSNR, and SD:} Images generated using X-Fork are better than other methods in terms of SSIM, PSNR and SD. We find that X-Fork improves over Zhai \textit{et al.} by 5.03\% in SSIM, 8.93\% in PSNR, and 12.35\% in SD.

\noindent \textbf{SVA dataset.} The quantitative evaluation on SVA dataset is illustrated in Table \ref{tab:sva_quant}.

Inception Score: H-Regions generates images that have the inception score closest to that of real data.
% among the nine different methods.

Accuracy: 
H-Seq method performs best in terms of accuracy.

KL(model $\|$ data):
Images synthesized using X-Seq method have the closest distribution to the ground truth distribution among all the methods.

{FID Score:} The H-Regions performs the best in terms of FID score in SVA dataset. The H-models perform better than their X- counterparts.

{SSIM, PSNR, and SD:}
X-Seq achieves the highest numbers in terms of SSIM and PSNR whereas H-Regions has the best SD.

Also, H-Pix2pix already performs very good compared to X-Pix2pix because homography simplified the learning task by transforming the image to the target view. H- methods outperform their X- counterparts for most of the evaluation metrics.

Because there is no consensus in evaluation of GANs, we had to use several scores. \cite{Theis2016a} show that these scores often do not agree with each other and this was observed in our evaluations as well. %So, it is difficult to infer which of the methods does the best. 
Nonetheless, we find that the proposed methods are consistently superior to the baselines in terms of quantitative and qualitative evaluations.

\begin{table}[t]
 \small
  \centering
  \renewcommand{\tabcolsep}{.2mm}  
  
  \caption{\small KL Divergence between model and data distributions on {\bf Dayton} and {\bf CVUSA} datasets.}
  \vspace{-10pt}
  \label{tab:model_data_KL}
    \begin{tabular}{llccc}
        \toprule
       \textbf{Dir.} & \textbf{Method} & \textbf{Dayton(64$\times$64)} & \textbf{Dayton(256$\times$256)} & \textbf{CVUSA}\\
        \midrule
        & \cite{zhai2017crossview} & \textbf{--} & \textbf{--} & $ 27.43 \pm 1.63 $\\
        & X-Pix2pix & $ 6.29 \pm  0.8 $ & $ 38.26 \pm 1.88 $ &   $ 59.81 \pm 2.12 $\\
    a2g & X-SO &  $4.99 \pm 0.71$  & $7.20 \pm 1.37 $ &   $414.25 \pm 2.37$\\  
 & X-Fork &   $ \textbf{3.42} \pm \textbf{0.72} $ & $6.00 \pm 1.28$ &  $ \textbf{11.71} \pm \textbf{1.55} $
\\
    & X-Seq & $ 6.22 \pm 0.87 $ & $ \textbf{5.93} \pm  \textbf{1.32}$ & $ {15.52} \pm {1.73} $\\
    \midrule 
    & X-Pix2pix & $ 6.39 \pm  0.90 $ & $ 7.88 \pm 1.24 $ &   \textbf{--} \\
         & X-SO &  $16.45 \pm 0.9 $ & $ 9.15 \pm 1.22$ &   --\\  

    g2a & X-Fork &   $ \textbf{4.45} \pm \textbf{0.84} $ & $\textbf{6.92} \pm \textbf{1.15} $ & \textbf{--}
\\
    & X-Seq & $ 7.20 \pm 0.92 $ & $ {7.07} \pm  {1.19}$ & \textbf{--}\\
        \bottomrule
        % \vspace{-10pt}
  
    \end{tabular}
\end{table}

\begin{table}[t]
 \small
  \centering
  \renewcommand{\arraystretch}{.6}
  \renewcommand{\tabcolsep}{.45mm}  
  
  \caption{\small Frechet Inception Distance (FID) scores on {\bf Dayton} and {\bf CVUSA} datasets.}
  \vspace{-10pt}
  \label{tab:fid}
    \begin{tabular}{llccc}
        \toprule
       \textbf{Dir.} & \textbf{Method} & \textbf{Dayton(64$\times$64)} & \textbf{Dayton(256$\times$256)} & \textbf{CVUSA}\\
        \midrule
        & \cite{zhai2017crossview} & \textbf{--} & \textbf{--} & $ 571.32 $\\
        & X-Pix2pix & $ 361.10 $ & $ 64.97 $ &   $ 644.45 $\\
    a2g & X-SO &  724.42  & 94.56 & 891.62   \\  
 & X-Fork &   \textbf{227.16} & $ 71.47 $ &  $ 185.42 $
\\
    & X-Seq & $ 676.14 $ & \textbf{46.34} & \textbf{89.12} \\
    \midrule 
    & X-Pix2pix & $ 323.54 $ & $ 71.35 $ &   \textbf{--} \\
         & X-SO & 412.38 & 108.14 &   --\\  

    g2a & X-Fork &   \textbf{239.94} & $ 80.03 $ & \textbf{--}
\\
    & X-Seq & $ 388.85 $ & \textbf{69.73} & \textbf{--}\\
        \bottomrule
        % \vspace{-10pt}
  
    \end{tabular}
    % \vspace{-5pt}
\end{table}

\begin{table*}[htbp]
 \small
   \renewcommand{\arraystretch}{.8}
  \centering
%   \vspace{-10pt}
  \caption{\small SSIM, PSNR and Sharpness Difference between real data and samples generated using different methods on \textbf{Dayton} and \textbf{CVUSA} datasets.}
  \vspace{-10pt}
  \label{tab:ssim_psnr_sd}
  
    \begin{tabular*}{\textwidth}{l @{\extracolsep{\fill}} lcccccccccc}
        \toprule
             \multicolumn{1}{l}{\textbf{Dir.}} &   \multicolumn{1}{l}{\textbf{Methods}} & \multicolumn{3}{c}{\textbf{Dayton (64 $\times$ 64)}} & \multicolumn{3}{c}{\textbf{Dayton (256 $\times $256)}} & \multicolumn{3}{c}{\textbf{CVUSA}}     \\
                \cmidrule(lr){3-5}
\cmidrule(lr){6-8} \cmidrule(lr){9-11}
    $\rightleftarrows$     &  & SSIM $\uparrow$ & PSNR $\uparrow$ & SD $\downarrow$ & SSIM  & PSNR & SD & SSIM & PSNR & SD \\
        \midrule
   &  \cite{zhai2017crossview} & \textbf{--} & \textbf{--} & \textbf{--} & \textbf{--} & \textbf{--} & \textbf{--}  & 0.4147 & 17.4886 & 16.6184 \\
    & X-Pix2pix  & 0.4808 & 19.4919 & 16.4489 & 0.4180 & 17.6291 & 19.2821  & 0.3923 & 17.6578 & 18.5239 \\
     a2g &  X-SO & 0.4960 & 19.7442 & 16.6670 & 0.4772 & 19.6203 & 19.2939 & {0.3451} & {17.6201} & {16.9919} \\

   & X-Fork & 0.4921 & 19.6273 & 16.4928 & 0.4963 & 19.8928 & 19.4533  & \textbf{0.4356} & \textbf{19.0509} & \textbf{18.6706}\\
   & X-Seq & \textbf{0.5171} & \textbf{20.1049} & \textbf{16.6836} & \textbf{0.5031} & \textbf{20.2803} & \textbf{19.5258}  & 0.4231 & 18.8067 & 18.4378 \\
    \midrule
  &  X-Pix2pix & 0.3675 & 20.5135 & {14.7813} & 0.2693 & 20.2177 & 16.9477 & \textbf{--} & \textbf{--} & \textbf{--} \\
  &  X-SO & 0.3254 & 16.5433 & \textbf{16.2329} & 0.2620 & 19.9827 & 16.8748 & \textbf{--} & \textbf{--} & \textbf{--} \\
  g2a &  X-Fork & \textbf{0.3682} & \textbf{20.6933} & {14.7984} & \textbf{0.2763} & \textbf{20.5978} & \textbf{16.9962}  &  \textbf{--} & \textbf{--} & \textbf{--}\\
 &   X-Seq & {0.3663} & {20.4239} & {14.7657} & {0.2725} & {20.2925} & {16.9285}  & \textbf{--} & \textbf{--} & \textbf{--} \\
        \bottomrule
         \vspace{-10pt}
    \end{tabular*}
\end{table*}  

% SSIM: 0.47710365814447	
% PSNR: 19.672570469762	
% Sharpness: 19.319654591607		

\begin{table*}[htbp]
 \small
   \renewcommand{\arraystretch}{.8}
  \centering
%   \vspace{-15pt}
  \caption{\small Quantitative evaluation of samples generated using different methods on {\bf SVA} Dataset in a2g direction.}
  \vspace{-10pt}
  \label{tab:sva_quant}
  
    \begin{tabular*}{\textwidth}{l @{\extracolsep{\fill}} cccc|cccc|c}
        \toprule  
 \textbf{Methods} & X-Pix2pix & X-SO & X-Fork & X-Seq & H-Pix2pix & H-SO & H-Fork & H-Seq & H-Regions \\
     \midrule
*Inception Score, all $\uparrow$ & {2.0131} & {2.4951} & {2.1888} & {2.2232}  & {2.1906} & 2.3202 & {2.3202}  & 2.2394 & \textbf{2.6328} \\
*Inception Score, Top-1 $\uparrow$ & {1.7221} & {1.8940} & {1.9776} & {1.9842}  & {1.9507} & 1.9410 & {1.9525}  & 1.9892  & \textbf{2.0732} \\
*Inception Score, Top-5 $\uparrow$ & {2.2370} & {2.6634} & {2.3664} & {2.4344}  & {2.4069} & 2.7340  & {2.3918}  & 2.4385 & \textbf{2.8347} \\

          \midrule
Accuracy (Top-1, all) $\uparrow$ & {8.5961} & {7.5146} & {17.3794} & {19.5056}  & {18.0706} & 5.2444 & {18.0182}  & \textbf{20.7391}  & 15.4803 \\ 
 Accuracy (Top-1, 0.5) $\uparrow$ & {30.3288} & {30.9507} & {53.4725} & {57.1010}  & {54.8068} & 26.4697 & {51.0756}  & \textbf{57.5378}  & 48.0767 \\ 
Accuracy (Top-5, all) $\uparrow$ & {9.0260} & {10.3905} & {23.8315} & {25.8807}  & {23.4400} & 5.2544  & {26.6746}  & \textbf{28.5517} & 21.8225 \\ 
 Accuracy (Top-5, 0.5) $\uparrow$ & {29.9102} & {38.9822} & {63.5045} & {65.3005}  & {62.3072} & 31.9527  & {62.8166}  & \textbf{67.4649} & 56.8994 \\ 
 
          \midrule
KL(model  $\mid$$\mid$ data) $\downarrow$ & {19.5553}  & {12.0906} & {4.1925} & \textbf{3.7585} & {4.2894} & 12.8761 & {4.7246}  & 4.4260  & 6.0638 \\ 

\midrule
SSIM $\uparrow$ & {0.3206} & {0.4552} & {0.4235} & \textbf{0.4638}  & {0.4327} & 0.4457 & {0.424}  & 0.4249 & 0.4044 \\
PSNR $\uparrow$ & {17.9944} & 21.5312 & {21.24} & \textbf{22.3411}  & 21.686 & 21.7709 & 21.6327  & 21.4770  & 20.9848 \\
SD $\downarrow$ & {17.0254} & 17.5285 & 16.9371 & {17.4138}  & 16.9468 & {17.3876} & 16.8653  & 17.5616 & \textbf{17.6858}\\
\midrule
FID Score $\downarrow$ & {859.66} & {443.79} & {129.16} & {118.70}  & {117.13} & {1452.88} &  {109.43} & 95.12 & \textbf{88.78}  \\

        \bottomrule
%         \vspace{-20pt}
    \end{tabular*}
{*Inception Score for real (ground truth) data is 3.0347, 2.3886 and 3.3446 for all, top-1 and top-5 setups respectively.}
\vspace{-10pt}
\end{table*}

\vspace{-5pt}
\section{Discussion and Conclusion}  
\label{sec:conclusion}
We explored image generation using conditional GANs between two drastically different views. Generating semantic segmentations together with images in the target view helps the networks learn better images compared to the baselines. Using homography to guide the cross-view synthesis allows preserving the overlapping regions between the views. Extensive qualitative and quantitative evaluations testify the effectiveness of our methods. Future research can explore other cues such as edge maps to facilitate synthesis tasks. Also, efforts can be put towards automatically finding regions that we defined manually. The challenging nature of the problem leaves room for further improvements. Potential application of this work can be in bridging the big domain-gap between street-view and aerial imagery for cross-view image geo-localization, multi-view object synthesis and so on.

\bibliographystyle{model2-names}
\bibliography{refs}

\end{document}